\documentclass[lettersize,journal]{IEEEtran}  

\IEEEoverridecommandlockouts                              

\usepackage{amsmath,amssymb,amsfonts}
\usepackage{algorithm}
\usepackage{algorithmic}
\usepackage{graphicx}
\usepackage{textcomp}
\usepackage{xcolor}
\usepackage{multirow}
\usepackage{booktabs}
\usepackage{bm}
\usepackage{mathrsfs}
\usepackage{float}
\usepackage{subfigure}
\usepackage{marvosym}

\title{\LARGE \bf
Never Too Cocky to Cooperate: An FIM and RL-based USV-AUV Collaborative System for Underwater Tasks in Extreme Sea Conditions
}

\author{Jingzehua Xu$^{1}$$^\dagger$, \IEEEmembership{Student Member, IEEE}, Guanwen Xie$^{1}$$^\dagger$, \IEEEmembership{Student Member, IEEE}, Jiwei Tang$^{2}$,\\ Yimian Ding$^{1}$, \IEEEmembership{Student Member, IEEE}, Weiyi Liu$^{1}$, \IEEEmembership{Student Member, IEEE}, Junhao Huang$^{3}$,\\ Shuai Zhang$^{4}$, \IEEEmembership{Member, IEEE}, and Yi Li$^{1,}\textsuperscript{\Letter}$, \IEEEmembership{Member, IEEE}
\thanks{This article has be presented in part at the IEEE International Conference on Acoustics, Speech and Signal Processing (ICASSP), Hyderabad, India, in April 2025.}
\thanks{$^{1}$J. Xu, G. Xie, Y. Ding, W. Liu and Y. Li are with Tsinghua Shenzhen International Graduate School, Tsinghua University, Shenzhen, 518055, China. E-mail: \{xjzh23, xgw24, dingym24, liuwy24\}@mails.tsinghua.edu.cn, liyi@sz.tsinghua.edu.cn.}%
\thanks{$^{2}$J. Tang is with Department of Data and Systems Engineering, The University of Hong Kong, Pokfulam, Hong Kong, E-mail: tangjiwei7@gmail.com.}
\thanks{$^{3}$J. Huang is with School of Engineering Science, University of Chinese Academy of Sciences, Beijing, China, E-mail: huangjunhao22@mails.ucas.ac.cn.}
\thanks{$^{4}$S. Zhang is with Department of Data Science, New Jersey Institute of Technology, NJ 07102, USA. E-mail: sz457@njit.edu.}
\thanks{$^\dagger$ These authors contributed equally to this work.}
\thanks{$\textsuperscript{\Letter}$ Corresponding author.}
}
\begin{document}

\markboth{IEEE TRANSACTIONS ON MOBILE COMPUTING}%
{Shell \MakeLowercase{\textit{et al.}}: A Sample Article Using IEEEtran.cls for IEEE Journals}

\maketitle

\begin{abstract}
This paper develops a novel unmanned surface vehicle (USV)–autonomous underwater vehicle (AUV) collaborative system designed to enhance underwater task performance in extreme sea conditions. The system integrates a dual strategy: (1) high-precision multi-AUV localization enabled by Fisher Information Matrix (FIM)-optimized USV path planning, and (2) a Reinforcement Learning (RL)-based cooperative planning and control framework for multi-AUV task execution.  Extensive experimental evaluations in the underwater data collection task demonstrate the system’s operational feasibility, with quantitative results showing significant performance improvements over baseline methods. The proposed system exhibits robust coordination capabilities between USV and AUVs while maintaining stability in extreme sea conditions.  To facilitate reproducibility and community advancement, we provide an open-source simulation toolkit available at: https://github.com/360ZMEM/USV-AUV-colab .
\end{abstract}


\section{Introduction}

Autonomous underwater vehicles (AUVs) have become indispensable tools for underwater exploration, enabling critical tasks such as environmental monitoring \cite{1}, seabed mapping \cite{2}, and biological research \cite{3}. However, traditional approaches relying on single-AUV systems or loosely coordinated swarms face significant limitations in positioning accuracy, scalability, and energy efficiency \cite{4}, particularly in extreme sea conditions where dynamic currents and turbulence exacerbate these challenges \cite{5}.

A central bottleneck lies in AUV localization: conventional methods like inertial navigation systems (INS) and standalone acoustic beacons suffer from error accumulation and signal degradation in dynamic environments, leading to unreliable positioning over prolonged missions \cite{6}. To address this, integrating AUVs with an unmanned surface vehicle (USV) as a mobile base station offers a promising solution \cite{7}. The USV provides real-time, high-precision positioning by acting as a communication and navigation hub, delivering GPS-corrected acoustic updates to submerged AUVs \cite{8}. This synergy not only mitigates error drift but also enhances multi-AUV coordination \cite{9}, making USV-AUV collaboration a growing research focus.

\begin{figure}[!t]
    \centering
    \includegraphics[width=1.0\linewidth]{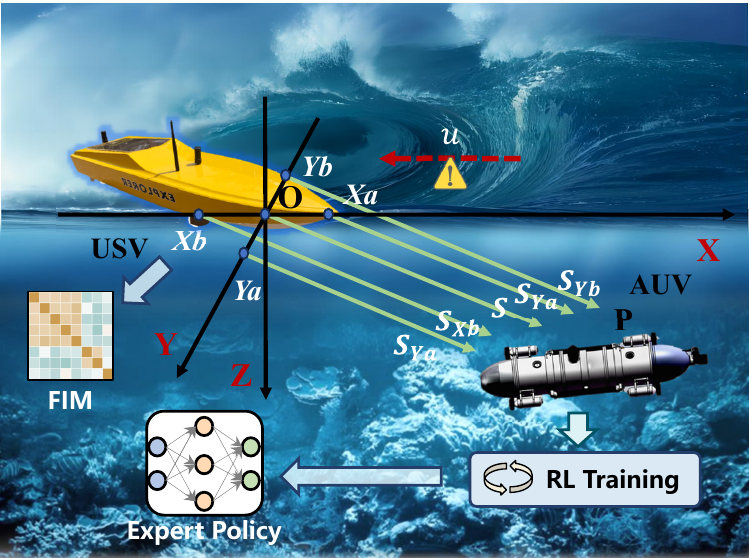}
    \caption{\textbf{The proposed USV-AUV collaborative system is designed to perform underwater tasks efficiently.} It employs a dual-strategy approach, combining FIM-based USV path planning for precise AUV positioning and RL-based training to enhance adaptive decision-making in AUVs. This integrated system ensures effective cooperation between USV and AUVs, optimizing mission performance in extreme sea conditions.}
    \label{fig:system}
\end{figure}

Despite recent advances, existing USV-AUV frameworks often lack the intelligence and adaptability required for complex tasks in extreme sea conditions \cite{10}. For instance, while Ultra-Short Baseline (USBL) positioning via USV improves localization, most approaches either fix the USV at static coordinates or employ heuristic path planning \cite{11}. Such rigidity results in inefficient USV-AUV linkage and fails to adapt to dynamic ocean environments, leading to suboptimal positioning and task performance. Further, traditional AUV control strategies (e.g., PID, fuzzy control) rely heavily on precise dynamic models, exhibit poor adaptability to turbulence, and struggle with multi-objective optimization or high-dimensional sensor data \cite{12}, which has motivated the application of learning-based methods. Crucially, many studies treat ocean dynamics as negligible noise, which is an unrealistic assumption in extreme conditions \cite{13}.

To address these challenges, a dual strategy is essential: an adaptive planning and control policy to enable dynamic decision-making in uncertain environments without relying on rigid models while optimizing multi-AUV collaboration for efficient task execution, and an uncertainty- and information-aware path planning method to improve the accuracy of AUV positioning via USV to enhance the robustness of the USV-AUV system, mitigating localization errors caused by extreme sea conditions.

This motivates the integration of reinforcement learning (RL) and Fisher Information Matrix (FIM) optimization, a synergistic approach that jointly tackles control and localization challenges \cite{14}. On one hand, RL’s model-free adaptive optimization enables AUVs to learn robust policies through environmental interaction, overcoming nonlinearities and time-varying disturbances without precise dynamic models \cite{15,16}. Multi-Agent RL (MARL) further enhances cooperative planning and control, enabling efficient multi-AUV coordination under communication constraints \cite{17}. On the other hand, FIM-based optimization directly quantifies and minimizes positioning uncertainty via USV path planning to maximize the system’s information content, thereby reducing the Cramér-Rao Bound (CRB) for AUV navigation \cite{14}. By combining RL’s adaptability with FIM’s precision, the proposed system achieves both intelligent decision-making and high-accuracy localization, which is a critical requirement for extreme sea scenarios.

Based on the above analysis, this paper develops a FIM and RL-based USV-AUV collaborative system for underwater tasks in extreme sea conditions to improve mission completion efficiency. The contributions are listed as follows:

\begin{itemize}
\item A novel USV-AUV collaborative system is developed, which integrates a FIM- and RL-based dual strategy to enable dynamic
decision-making and uncertainty- and information-aware capabilities to improve mission completion efficiency for underwater tasks. 

\item This system achieves accurate positioning of AUVs via USV path planning by minimizing the determinant of the system's FIM. Based on this, by integrating environment-aware abilities
into the state space and employing a shaped reward
function for multi-AUV collaboration, we further utilize RL to empower the multi-AUV with intelligence
and adaptability for the underwater task.

\item Through comprehensive experiments in the underwater data collection task, our system demonstrates superior feasibility and excellent performance, exhibiting robustness under extreme sea conditions.
\end{itemize}

The remainder of this paper is organized as follows: Section II reviews related work on USV-AUV collaboration in underwater tasks. Section III details the methodology of this work. Section IV describes the environmental simulations and experimental results. Finally, Section V concludes the paper and outlines future directions. For ease of reference, the main symbols and their corresponding explanations used throughout this paper are summarized in Table I.

\begin{table}[!t]
\centering
\caption{Main Symbols and Explanations}
\label{tab:symbols_concise}
\setlength{\tabcolsep}{2.0mm}{
\begin{tabular}{c|c}
\hline
{\bf Symbols} & {\bf Definition}\\
\hline
\hline
$\mathbf{p}_{\text{target}}^k$ & Target SN position for AUV $k$\\
\hline
$\mathbf{Z}_k$ & USBL measurement for AUV $k$\\
\hline
$\mathbf{h}_k(\mathbf{X})$ & Measurement model\\
\hline
$\Delta\varphi_{x,k},\Delta\varphi_{y,k}$ & Phase differences (x/y)\\
\hline
$f,~c,~d$ & Frequency; sound speed; array spacing\\
\hline
$S_k$ & Slant range USV$\leftrightarrow$AUV $k$\\
\hline
$\gamma_k,~\varphi_k$ & Elevation; azimuth of AUV $k$\\
\hline
$A_k$ & Geometric weight\\
\hline
$\alpha_{ij}=\varphi_i-\varphi_j$ & Azimuth separation\\
\hline
$S_0,~\gamma_0$ & Slant/elevation (symmetric case)\\
\hline
$\chi=\sum_{i<j}\sin^2\alpha_{ij}$ & Angular-diversity factor\\
\hline
$\mathcal{S},\mathcal{A},\mathcal{R},\mathcal{P},\gamma$ & State, action, reward, dynamics, discount\\
\hline
$\pi_\phi$ & Deterministic policy (actor)\\
\hline
$Q_{\theta_1},Q_{\theta_2}$ & Twin critics (Q-functions)\\
\hline
$\phi',~\theta_i'$ & Target networks\\
\hline
$\lambda_Q,~\lambda_\pi$ & Critic/actor learning rates\\
\hline
$\tau$ & Soft-update rate\\
\hline
$\mathcal{D}$ & Replay buffer\\
\hline
$\rho^\pi(s)$ & State visitation\\
\hline
$s_t^k$ & State of AUV $k$ at $t$\\
\hline
$\Delta\mathbf{p}_j^k$ & Rel. pos. to AUV $j$ (norm.)\\
\hline
$\Delta\mathbf{p}_{\text{target}}^k$ & Rel. pos. to target SN (norm.)\\
\hline
$\tilde{\mathbf{p}}_t^k$ & Self-position (norm.)\\
\hline
$\rho_{\text{overflow}}$ & Data-overflow ratio\\
\hline
$\mathbb{I}_{\text{border}}^k$ & Border flag\\
\hline
$a_t^k=[v_{\text{norm}},\theta_{\text{norm}}]^\top$ & Normalized speed/heading\\
\hline
$d_k^{\text{target}}$ & AUV–target distance\\
\hline
$d_{kj}$ & Inter-AUV distance\\
\hline
$N_{\text{DO}}$ & Overflow count\\
\hline
$N_{\text{POI}}$ & Total number of the underwater sensor nodes\\
\hline
$\mathbb{I}_{\text{TL}}$ & Transmission-success flag\\
\hline
$E_k$ & Energy of AUV $k$\\
\hline
$u,v$ & Wave velocities ($x',y'$)\\
\hline
$\eta$ & Free-surface elevation\\
\hline
$\Delta t,\Delta x,\Delta y$ & Time/space steps (FDM)\\
\hline
$V_{x,i,j},V_{y,i,j}$ & Turbulent velocity (grid)\\
\hline
$\varpi_{i,j}$ & Vorticity (grid)\\
\hline
$(x_0,y_0)$ & Vortex center\\
\hline
$\Gamma,~\delta$ & Vortex intensity; core radius\\
\hline
\end{tabular}}
\end{table}

\section{Related Work}
\subsection{Advances in USV-AUV Collaborative Systems}
The USV-AUV collaborative system has garnered significant attention due to its potential in marine exploration and environmental monitoring. Early research primarily addressed communication and coordination challenges, including acoustic communication modems and surface relay systems \cite{18}, as well as task allocation mechanisms in which USVs serve as mobile base stations \cite{19}. Concurrently, cooperative navigation frameworks developed for GPS-denied environments significantly enhance AUV positioning capabilities \cite{20}.

Recent studies have focused on multi-agent autonomous collaboration, with machine learning algorithms enabling cooperative tasks such as seabed mapping \cite{21}. Hybrid communication systems combining acoustic and radio links have improved system reliability \cite{22}. Although challenges in real-time decision-making under environmental uncertainties persist, emerging technologies such as swarm intelligence and RL offer solutions for scalable collaboration \cite{23}.

\subsection{FIM-based Approaches for USV}
Recent studies have extensively applied the FIM to enhance the efficiency and robustness of USV and AUV operations in dynamic marine environments. The foundational work by Wilson \textit{et al}. \cite{25} first formulated trajectory optimization via FIM maximization to improve parameter observability, establishing the theoretical groundwork for information-driven mobile robot planning. Building on this, Bahr \textit{et al}. \cite{7} demonstrated the use of mobile surface beacons for cooperative underwater navigation, aligning with the concept of a dynamic USV acting as a moving information source. Subsequently, Wang \textit{et al}. \cite{14} integrated USBL measurements with FIM optimization to derive Dubins-path-based USV trajectories that directly improve AUV localization precision, while Paull \textit{et al}. \cite{18} provided a comprehensive review tracing the evolution from traditional acoustic navigation to information-theoretic trajectory optimization.

In practical applications, FIM has also been widely adopted in adaptive sampling strategies, where USV trajectories are optimized to enhance environmental field reconstruction (e.g., salinity or temperature mapping) \cite{24}. These approaches often combine FIM metrics with gradient-based or sampling-based planners to balance exploration efficiency and motion constraints \cite{25}. More recent efforts have extended FIM to real-time USV path planning under environmental disturbances and communication limitations, as well as to cooperative USV–AUV systems that maximize the joint FIM determinant through distributed optimization \cite{14}. Despite these advances, challenges remain in computational scalability for high-dimensional FIM calculations and in maintaining robustness under sensor noise and model mismatches. Emerging solutions include Gaussian process-based FIM approximations \cite{28} and hybrid learning-control architectures that adaptively tune information-driven objectives \cite{29}.

Building upon these theoretical and practical foundations, our work extends FIM-based trajectory optimization into a multi-agent coordination framework, where the FIM explicitly models how the USV’s sensing and motion contribute to AUV localization accuracy in real time. This tight integration of information-theoretic modeling with RL-based control establishes a principled bridge between estimation theory and cooperative autonomy, constituting a key methodological innovation of our system.

\subsection{RL-based AUV Control and Planning}
RL has become a prominent approach for AUV control and planning in complex marine environments. Early studies employed value-based methods such as Q-learning for basic navigation and station-keeping tasks \cite{30}, while subsequent work adopted policy gradient algorithms to cope with dynamic ocean currents and uncertain disturbances \cite{31}. With the advent of RL, end-to-end control frameworks emerged that directly learn from raw sensor data \cite{32}, greatly improving adaptability but still facing challenges in partial observability, sample inefficiency, and safety assurance. To overcome these limitations, recent research has shifted toward hybrid, model-based, and cooperative RL frameworks that incorporate environmental priors and physical constraints. For example, Zhang \textit{et al}. \cite{38} introduced a hybrid control architecture combining RL with model-based control to enhance robustness and stability, while Wang \textit{et al}. \cite{39} integrated Model Predictive Control (MPC) and Control Barrier Functions (CBF) within an RL framework to adaptively tune control weights, achieving reliable trajectory tracking and obstacle avoidance in complex underwater conditions. Similarly, Xi \textit{et al}. \cite{40} proposed the Comprehensive Ocean Information D3QN (COID), which exploits real oceanographic data and a double dueling deep Q-network to enable flexible and data-driven path planning across dynamic marine environments.

In parallel, meta-learning and hierarchical mechanisms have been introduced to improve sample efficiency and generalization in non-stationary settings, as demonstrated by Fan \textit{et al}. \cite{41}, while multi-agent cooperative RL allows teams of AUVs to coordinate complex missions under communication constraints \cite{33}. Model-based RL approaches have also leveraged ocean simulation and real data assimilation to accelerate convergence and improve safety guarantees \cite{34,35}. Furthermore, Wozniak \textit{et al}. \cite{42} applied the twin-delayed deep deterministic policy gradient algorithm to control hydrobatic AUVs, demonstrating robust depth and pitch regulation in simulations and revealing the potential of RL-based controllers for agile underwater maneuvers, despite sim-to-real challenges.

Collectively, these studies demonstrate that RL-based AUV control and planning have evolved from simple navigation and motion control to multi-objective, cooperative, and constraint-aware optimization, forming a solid foundation for adaptive underwater autonomy. However, most existing RL frameworks still focus on single-vehicle control under simplified conditions, lacking treatment of multi-AUV cooperation and localization coupling. To address these gaps, our work integrates information-driven positioning and cooperative RL within a unified framework. Specifically, the USV optimizes its trajectory to maximize the FIM, improving AUV localization accuracy, while the resulting high-fidelity positional information is embedded into the multi-agent RL loop to enhance coordination and situational awareness. Moreover, the USV–AUV localization relationship is mathematically formulated through the FIM, enabling decoupled yet interactive learning between localization and policy optimization. This explicit information-coupled design offers interpretability, modularity, and theoretical rigor, marking a key methodological novelty that bridges information-theoretic modeling with reinforcement learning-based multi-AUV coordination in complex marine environments.

\begin{figure*}[!t]
    \centering
    \includegraphics[width=1.0\linewidth]{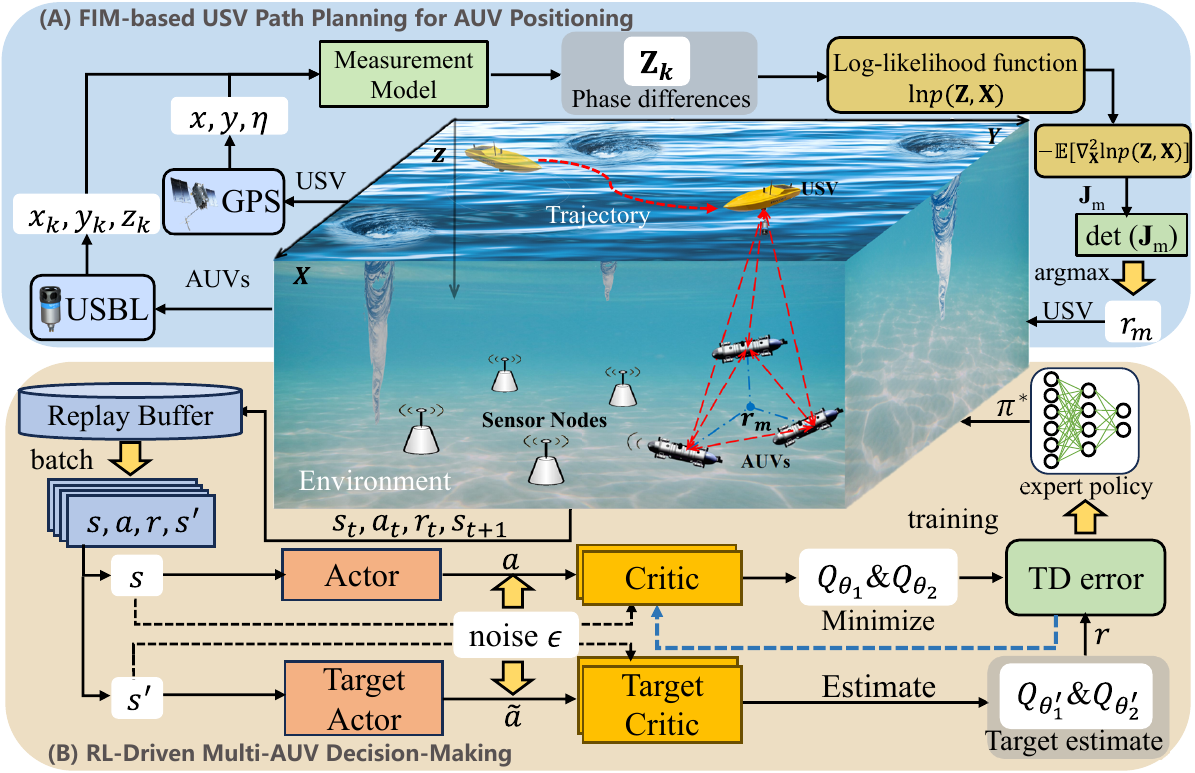}
    \caption{\textbf{The overall architecture of proposed USV-AUV collaborative system,} which integrates FIM optimization and RL to enhance adaptability and precision in dynamic marine environments. Through optimization of FIM's determinant, USV can realize path planning to improve AUV positioning accuracy. Besides, TD3 algorithm enables multi-AUV systems to learn efficient navigation and collaboration strategies through iterations of RL training. Together, they ensure robust performance despite disturbances like tides or turbulence, continuously refining behavior through real-time feedback in the extreme sea conditions.}
    \label{fig:2}
\end{figure*}

\section{Methodology}
\subsection{System Architecture}
The proposed USV-AUV collaborative system architecture combines FIM optimization and RL training to achieve dynamic decision-making, uncertainty-aware capabilities, and adaptive behavior in challenging marine environments.  As illustrated in Fig. 2, the system adopts a dual-strategy framework: FIM optimizes AUV positioning accuracy through USV path planning, while RL enhances multi-AUV intelligence and adaptability to dynamic ocean conditions.

Specifically, the system employs FIM to maximize AUV localization precision by minimizing the determinant of the FIM, thereby reducing estimation uncertainty in the underwater environment. The USV dynamically adjusts its path planning based on real-time FIM calculations, ensuring optimal positioning support for the AUVs. This mathematical approach enhances system robustness, even when external disturbances such as tidal waves or turbulence degrade sensor measurements.

To further improve adaptability, the system integrates an RL framework in which the state space incorporates environment-aware data, and the reward function evaluates the USV-AUV collaboration efficiency. This enables multi-AUV systems to learn and refine navigation and task execution strategies under extreme conditions, thereby increasing mission completion efficiency. By continuously adapting to real-time environmental feedback, the RL component ensures sustained performance optimization.

\subsection{FIM-based USV Path Planning for AUV Positioning}
Our system enables accurate underwater vehicle positioning through an optimized USV navigation strategy that leverages USBL technology while further utilizing the optimization of the FIM's determinant to achieve efficient USV path planning, thereby minimizing AUV positioning errors. As shown in Fig. 1 and Fig. 2, the system architecture consists of:
\begin{itemize}

\item A surface USV located at coordinates $(x, y, \eta)$ equipped with a USBL transducer array (inter-element spacing $d/2$) while using GPS for self-positioning.

\item Multiple AUVs deployed at underwater positions $(x_k, y_k, z_k) ,k=1,2,...,m$.
\end{itemize}

The positioning framework is built upon a measurement model in USBL that correlates observed phase differences with AUV locations. This relationship is mathematically expressed as \cite{USBL}:

\begin{equation}
    \mathbf{Z}_k = \mathbf{h}_k(\mathbf{X}) + \mathbf{u}_k, \quad \mathbf{X} = [x, y]^\top,
    \label{eq:measurement}
\end{equation}
where $\mathbf{h}_k = [\Delta\varphi_{x,k}, \Delta\varphi_{y,k}]^\top$ is the phase differences vector, whose elements can be expressed as:
    \begin{align}
        \Delta\varphi_{x,k} &= \frac{2\pi f d (x_k - x)}{c S_k} \label{eq:phase_x}, \\
        \Delta\varphi_{y,k} &= \frac{2\pi f d (y_k - y)}{c S_k} \label{eq:phase_y},
    \end{align}
where $c$ represents the speed of sound and $f$ indicates the signal frequency. Additionally,

$S_k = \sqrt{(x_k - x)^2 + (y_k - y)^2 + z_k^2}$ is the slant range,

$\mathbf{u}_k \sim \mathcal{N}(0, \sigma^2\mathbf{I})$ represents measurement noise.

To rigorously evaluate the system's localization accuracy, we derive the FIM from this measurement model. The FIM provides a quantitative measure of position estimation precision by analyzing the information content in the USBL phase difference measurements.

The FIM is derived by calculating the expectation of the second partial derivatives of the log-likelihood function. This quantifies how much information the measurements carry about the unknown USV position $\mathbf{X}$:
Under the assumption that \(\mathbb{E}[\mathbf{Z}_k - \mathbf{h}_k(\mathbf{X})] = 0\), the FIM is expressed as \cite{FIM}:
\begin{align}
    \mathbf{J}_m = -\mathbb{E}\left[\nabla_{\mathbf{X}}^2 \ln p(\textbf{Z},\textbf{X})\right],
\end{align}
\begin{equation}
    \ln p(\mathbf{Z}, \mathbf{X}) \!\!=\!\! -\!\frac{1}{2} \!\sum_{k=1}^m [\mathbf{Z}_k \!-\! \mathbf{h}_k(\mathbf{X})]^\top \mathbf{R}^{-1} [\mathbf{Z}_k \!-\! \mathbf{h}_k(\mathbf{X})] \!+\! C,
    \label{eq:loglikelihood}
\end{equation}
where $\textbf{R}=\sigma^2\textbf{I}$ denotes the measurement noise covariance matrix, and $C$ is the constant related to \textbf{R}.

The determinant of $\mathbf{J}_m$ serves as a scalar metric for overall positioning accuracy. By simplifying $\det(\mathbf{J}_m)$ for geometric interpretation and optimization, we isolate two critical geometric factors:

\begin{equation}
    \det(\mathbf{J}_m) \!\propto\! \underbrace{(\sum_{k=1}^m \frac{1}{S_k^2}\sum_{k=1}^m \frac{{\rm sin^4}\gamma_k}{S_k^2})^2}_{\text{Depth and Slant Factor}} \!+\!\!\!\! \underbrace{\sum_{1 \leq i<j \leq m}^m \!\!\!\!\!A_iA_j\sin^2(\varphi_i\!\!-\!\!\varphi_j)}_{\text{Angular Diversity}},
    \label{eq:det_fim}
\end{equation}
where $A_k=\frac{p_k^4-2S^2_kp^2_k}{2S_k^6}$, with $p_k^2=x_k^2+y_k^2$, while ${\rm sin}\gamma_k\!\!=\!\!\frac{z_k}{S_k}$, and $\varphi_k = \arctan(y_k/x_k)$ denotes the azimuth angle. The first term emphasizes depth and slant-dependent observability, while the second term penalizes collinear AUV placements. This directly motivates the optimal formations, revealing that optimal positioning requires:
\begin{enumerate}
    \item Minimizing slant ranges $S_k$ to AUVs.
    \item Maximizing angular separation between AUV projections through $\sin^2(\varphi_i-\varphi_j)$ terms.
    \item Balancing depth-to-range ratios through $\sin^4\gamma_k$ terms.
\end{enumerate}

For the convenience of research without loss of generality, this work only considers the influence of the relative angles between AUVs and the target on positioning accuracy. When the distance between the AUVs and the USV is equal and the AUVs are at the
same depth, only the angle of the AUVs is considered, and the influence of AUV formation
on positioning accuracy is only related to ${\rm \!sin}^{2}\alpha_{ij}$:

\begin{equation}
    {\rm det}\!\left(\mathbf{\mathit{\textbf{{J}}}}_{m}\!\right) \!= \!\left(\!\right.\! \frac{4 \pi^{2} f^{2} d^{2}\!}{\sigma^{2} c^{2}} \left.\!\right)^{2} \!\!\left[3 m \frac{{\rm sin}^{2} \gamma_{0}}{S_{0}^{4}} \!+\! \frac{ \left({\rm sin}^{4} \gamma_{0} \!+\! 1 \right)^{2}}{S_{0}^{4}}\chi\right]\!,
\end{equation}
where ${\rm sin}\gamma_0\!\!=\!\!\frac{z_k}{S_0}$, while $\chi\!\!=\!\!\sum_{1\leq i \textless j\leq m}^m {\rm \!sin}^{2}\alpha_{ij}$, with $\alpha_{ij}\!\!=\!\!\varphi_i\!-\!\varphi_j$ representing the angle between the projections of two AUVs and the USV. 

As illustrated in Fig. 2, by maximizing the determinant ${\rm det}\!\left(\!\mathbf{{\boldsymbol{J}}}_{m}\!\right)$, we can determine the optimal horizontal distance $r_m$ between the USV and multiple AUVs:
\begin{equation}
    r_m = \rm argmax \left\{d e t \left(\mathbf{\mathit{\textbf{{J}}}}_{m}\right)\right\}.
\end{equation}

By calculating the waypoints using Eq. (8) with the maximum $ \rm d e t \left(\mathbf{{\textbf{{J}}}}_{m}\right)$, we can implement USV path planning to minimize AUV positioning errors through the above process.

\subsection{RL-Driven Multi-AUV Decision-Making}
Our approach leverages the Twin Delayed Deep Deterministic Policy Gradient (TD3) algorithm, which extends the Deep Deterministic Policy Gradient (DDPG) algorithm with three key improvements: (1) clipped double Q-learning to mitigate overestimation bias, (2) delayed policy updates to stabilize training, and (3) target policy smoothing to prevent overfitting to narrow peaks in the value function \cite{TD3}.  These characteristics make TD3 particularly suitable for underwater vehicle control, where precise continuous actions are required and the dynamics are highly coupled across degrees of freedom.

In continuous control tasks, we model the environment as a Markov Decision Process (MDP) for further RL training with:
\begin{equation}
    <\mathcal{S},\mathcal{A},\mathcal{R},\mathcal{P}(\cdot|s,a),\gamma>,
\end{equation}
where continuous state space $\mathcal{S}$, continuous action space $\mathcal{A}$, and reward function $r: \mathcal{S} \times \mathcal{A} \to \mathbb{R}$. $\mathcal{P}(\cdot|s,a)\sim(0,1)$ is the state transition probability. $\gamma \in[0,1]$ represents the discount factor.

Given the state space $\mathcal{S}$ and the action space $\mathcal{A}$, our objective in the underwater tasks is to obtain the expert policy by maximizing the discounted cumulative reward:
\begin{equation}
    \pi^*\!=\! {\rm argmax} \{J(\pi)\}\!\! =\! {\rm argmax}\{\mathbb{E}_{\pi}\!\! \left[ \sum_{t=0}^\infty \gamma^t r(s_t, a_t) \right]\!\},
\end{equation}
where $\pi: \mathcal{S} \to \mathcal{A}$ is a deterministic policy. The agent's objective is to find the optimal policy $\pi^*$ that maximizes the expected return, and $\gamma \in [0,1)$ is the discount factor. This infinite-horizon formulation requires careful treatment of the value function approximation.

The action-value function $Q^\pi(s,a)$ represents the expected return when taking action $a$ in state $s$ and following policy $\pi$ thereafter. It satisfies the fundamental Bellman equation:

\begin{equation}
    Q^\pi(s,a) = \mathbb{E}_{s' \sim P(\cdot|s,a)} \left[ r(s,a) + \gamma Q^\pi(s', \pi(s')) \right].
    \label{eq:bellman}
\end{equation}

This recursive relationship forms the basis for temporal difference learning. However, when combined with function approximation and off-policy updates, it can lead to value overestimation.

To mitigate overestimation bias, TD3 maintains two separate Q-function approximators $Q_{\theta_1}$ and $Q_{\theta_2}$ with independent parameters. 

\begin{algorithm}[!ht]
\caption{RL Training using TD3}
\begin{algorithmic}[1]
\STATE Initialize policy $\pi_\phi$ and critics $Q_{\theta_1}, Q_{\theta_2}$
\STATE Initialize target networks $\phi' \leftarrow \phi$, $\theta_i' \leftarrow \theta_i$
\STATE Initialize replay buffer $\mathcal{D}$
\FOR{each episode}
\STATE Observe initial state $s_0$
\FOR{each timestep}
\STATE Sample action $a_t = \pi_\phi(s_t) + \epsilon$, $\epsilon \sim \mathcal{N}(0,\sigma)$
\STATE Execute $a_t$, observe $s_{t+1}$ and $r_t$
\STATE Store $(s_t, a_t, r_t, s_{t+1})$ in $\mathcal{D}$
\IF{update time}
\STATE Sample batch $(s,a,r,s') \sim \mathcal{D}$
\STATE Compute target actions $        \tilde{a} \!=\! \pi_{\phi'}(s') \!+\! \text{clip}(\epsilon,\! -c, c)$
\STATE Update critics using $y \!= r + \gamma \min_{i=1,2} Q_{\theta_i'}(s', \tilde{a})$
\IF{actor update delay}
\STATE Calculate policy gradient:

\qquad $\nabla\!_\phi\! J\!(\phi) \!=\! \mathbb{E}[\nabla_a Q_{\theta_1}\!(s,a)|_{a=\!\pi_\phi(s)\!}\nabla_\phi \pi_\phi(s)]$
\STATE Q-networks update:

\STATE \qquad $\theta_i \leftarrow \theta_i - \lambda_Q \nabla_{\theta_i} \mathbb{E}[(Q_{\theta_i}(s,a) - y)^2]$

\STATE Delayed policy update:
\STATE \qquad $\phi \leftarrow \phi + \lambda_\pi \nabla_\phi J(\phi)$
\STATE Target network updates:
\STATE \qquad $\theta_i' \leftarrow \tau \theta_i + (1-\tau)\theta_i'$
\STATE \qquad $\phi' \leftarrow \tau \phi + (1-\tau)\phi'$
\ENDIF
\ENDIF
\ENDFOR
\ENDFOR
\end{algorithmic}
\end{algorithm}

The target noise $\epsilon$ serves as a regularizer, preventing the policy from exploiting overestimated Q-values. The clipped Gaussian noise ensures that the perturbation stays within reasonable bounds:

\begin{equation}
    \pi_{\text{target}}(s') = \pi_{\phi'}(s') + \text{clip}(\epsilon, -c, c).
    \label{eq:target_smoothing}
\end{equation}

This smoothing can be interpreted as adding a small entropy term to the deterministic policy.

The target value computation uses the minimum of these two estimates. Specifically, we maintain two Q-networks $Q_{\theta_1}$ and $Q_{\theta_2}$ with target networks:
\begin{equation}
    y = r + \gamma \min_{i=1,2} Q_{\theta_i'}(s', \pi_{\phi'}(s') + \text{clip}(\epsilon,\! -c, c)),
\end{equation}
where $\epsilon \sim \text{clip}(\mathcal{N}(0, \sigma), -c, c)$ adds noise to the target action for smoothing. This minimum operator provides a lower bound on the true value function, as we can show that:

\begin{equation}
    \min_{i=1,2} Q_{\theta_i'}(s',a') \leq Q^{\pi}(s',a'),
\end{equation}
for any $(s',a')$ under mild assumptions about approximation errors.

Thus, we can obtain that the Critic's loss function is based on the TD error of the Bellman equation:
\begin{equation}
    \mathcal{L}(\theta_i) = \mathbb{E}_{(s,a,r,s') \sim \mathcal{D}} \left[ (Q_{\theta_i}(s,a) - y)^2 \right], \quad i=1,2.
\end{equation}

The policy network updates less frequently than the Q-functions (typically every $d=2$ steps). This delay allows the value estimates to stabilize before the policy is adjusted, thereby reducing error propagation. According to the
deterministic policy gradient theorem
\begin{equation}
    \nabla_\phi J(\phi) = \int_{\mathcal{S}} \rho^\pi(s) \nabla_a Q^\pi(s,a)\big|_{a=\pi_\phi(s)} \nabla_\phi \pi_\phi(s) ds,
\end{equation}
we can obtain the policy gradient, which is computed as:
\begin{equation}
    \nabla_\phi J(\phi) = \mathbb{E}_{s \sim \mathcal{D}} \left[ \nabla_a Q_{\theta_1}(s,a)\big|_{a=\pi_\phi(s)} \nabla_\phi \pi_\phi(s) \right].
    \label{eq:policy_gradient}
\end{equation}

Note that only $Q_{\theta_1}$ is used for the policy gradient to prevent the "chasing" behavior that could occur if both Q-networks are used simultaneously.

In practice, we can estimate the policy gradient using the Monte Carlo method:
\begin{equation}
    \nabla_\phi J(\phi) \approx \frac{1}{N}\sum_{i=1}^N \nabla_a Q(s_i,a)\big|_{a=\pi_\phi(s_i)} \nabla_\phi \pi_\phi(s_i).
\end{equation}

Finally, we update the policy network, and update the target networks using soft update strategy. The overall RL training using the TD3 algorithm has been shown in pseudo-code Algorithm 1.

\section{Experiments and Analysis}
In this section, we verify the effectiveness of the proposed USV-AUV collaborative system in extreme sea conditions through comprehensive simulation experiments. Furthermore, we present the experimental results with further analysis and discussion.

    \begin{table}[!t]
        \centering
        \caption{\!Parameters of the Environment and Algorithm}
        \label{table:1}
        \begin{tabular}{lc} 
        \toprule
        Parameters & Values \\
        \midrule
        Experimental area    & 200m $\times$ 200m   \\
        USV sound level  & 135dB  \\ 
        USV transmit frequency & (12,14,16,18)khz (AUV1$\sim$4) \\
        Hydrophone array $d$ & 0.033m \\
        Water Depth & 120m \\ 
        Time step $\Delta t$  & 10s \\
        Space step $\Delta x$ & 4m \\ 
        Angular frequency $\omega$ & 2$\pi$/43200 rad/s \\ 
        Initial amplitude $\eta_0$ & 5m \\ 
        Training episode  & 450 \\ 
        Steps in each episode & 1000 \\ 
        Replay buffer size & $2*10^4$ \\ 
        Hidden layer size & $128$ \\
        Batch size & 64 \\
        Policy update interval & 2 \\
        Target policy noise & $\mathcal{N}(0,0.1)$, clipped to $\pm1.0$ \\
        Discount factor $\gamma$ & 0.97\\
        Actor learning rate & $1*10^{-3}$\\
        Critic learning rate & $1*10^{-3}$\\
        Soft update parameter &
        $1*10^{-3}$\\
        \bottomrule
        \end{tabular}
        \end{table}

    \subsection{Task Description and Experimental Settings} 
    \label{subsec:fc}
    Given the limited availability of open-source underwater tasks, we selected the underwater data collection task as a benchmark scenario to validate our system. In this application scenario, multiple AUVs and a USV work collaboratively to collect data from sensor nodes (SNs) deployed in the Internet of Underwater Things (IoUT) environments while simultaneously optimizing multiple objectives, including the serviced SN number (SSN) and sum data rate (SDR) maximization, collision avoidance, and energy consumption (EC) minimization. The specific parameter configurations employed in our study are comprehensively listed in Table I. For further technical specifications and detailed task parameters, readers may consult our prior research work \cite{5}.

As for the RL algorithm, the TD3 implementation uses separate networks for the actor (policy) and critics (Q-functions):

\begin{itemize}
    \item \textbf{Actor}: 3-layer MLP (128, 128, 128 units) with ReLU activations, outputting tanh-scaled actions.
    \item \textbf{Critics}: Twin 2-layer MLPs (128, 128 units) with ReLU activations.
\end{itemize}

The simulation experiments are conducted on a personal computer equipped with an AMD Ryzen 9 5950X processor and an NVIDIA GeForce RTX 4060 GPU, operating within a Python 3.8 virtual environment. Due to the adoption of an efficient computational approach for tidal wave simulation, our code required approximately 3 hours to complete 450 episodes under the aforementioned configuration, demonstrating low computational resource demands. For other parameters, please refer to Table II for a summary.

\subsection{Realistic Extreme Sea Condition Simulation}
The USV-AUV collaborative system's dual-domain operation (surface and subsurface) makes it particularly vulnerable to marine disturbances. To address this, we employ two-dimensional tidal wave equations and ocean turbulence models to realistically simulate extreme sea conditions, accurately replicating disturbances that impact AUV positioning and performance for rigorous system validation. The simulation effects of sea conditions at three different time points (25 s, 50 s, 75 s) are shown in Fig. 3.
    
If we denote the wave velocity as $\boldsymbol{V}\!_w\!=\![u,v]$, and the gravitational acceleration as $g$, the water level $\eta$ at coordinate point $(x^{'},y^{'}\!)$ can be calculated using the Finite Difference Method (FDM). The equations' forward difference scheme using FDM can be listed as follows \cite{36}:
\begin{subequations}
\begin{gather}
\frac{u_{i,j}^{n+1} - u_{i,j}^n}{\Delta t} + g \frac{\eta_{i+1,j}^n - \eta_{i,j}^n}{\Delta x} = 0, \\
\frac{v_{i,j}^{n+1} - v_{i,j}^n}{\Delta t} + g \frac{\eta_{i,j+1}^n - \eta_{i,j}^n}{\Delta y} = 0, \\
\frac{\eta_{i,j}^{n+1} \!\!-\! \eta_{i,j}^n}{\Delta t} \!\!+\!\! \frac{(u\! \cdot \! h)_{i,j}^n \!\!-\! (u\! \cdot \!h)_{i-1,j}^n}{\Delta x} \!\!+\!\! \frac{(v\! \cdot \!h)_{i,j}^n \!\!-\! (v\! \cdot\! h)_{i,j-1}^n}{\Delta y} \!=\! 0, 
\end{gather}
\end{subequations}
where $u_{i,j}^n$ and $v_{i,j}^n$ are the velocity components in $x'$ and $y'$ directions at grid point $(i,j)$ and time step $n$, $\eta_{i,j}^n$ is the water level height, $\Delta t$ is the time step, $\Delta x$ and $\Delta y$ are spatial steps, while $h$ represents the water depth.

Moreover, considering the vortex center position, the turbulent velocity and vorticity can be numerically computed at grid point $(i,j)$ through the ocean turbulence model's finite difference formulation \cite{37}:
\begin{subequations}
\begin{gather}
V_{x,i,j} \!\!=\!\! -\Gamma \!\! \cdot \! \frac{y_j \!\!-\! y_0}{2\pi \left[(x_i \!-\! x_0)^2 \!\!+\!\! (y_j \!\!-\!\!y_0)^2\right]} \!\!\left(\!\!1 \!\!-\!\! e\!^{-\frac{(x_i - x_0)^2 + (y_j - y_0)^2}{\delta^2}}\!\!\right)\!,\\
V_{y,i,j} \!=\! \Gamma \! \cdot \! \frac{x_i \!\!-\!\! x_0}{2\pi \left[(x_i \!\!-\!\! x_0)^2 \!+\! (y_j \!\!-\!\! y_0)^2\right]} \left(\!\!1 \!\!-\!\! e^{-\frac{(x_i - x_0)^2 + (y_j - y_0)^2}{\delta^2}}\!\!\right)\!, \\
\varpi_{i,j} = \frac{\Gamma}{\pi \delta^2} \cdot e^{-\frac{(x_i - x_0)^2 + (y_j - y_0)^2}{\delta^2}},
\end{gather}
\end{subequations}
where $\varpi$, $\delta$, and $\Gamma$ stand for the vorticity, radius, and intensity of the vortex, respectively.

\subsection{State, Action Space and Reward Function Design}
In this study, the full state vector of state $s_t^k \in \mathcal{S}$ for the $k$-th AUV at time $t$ is composed of:
\begin{equation}
    s_t^k \!\!=\!\! \left[\! \{\Delta \mathbf{p}_j^k\}_{j \neq k},\! \Delta \mathbf{p}_{\text{target}}^k,\! \tilde{\mathbf{p}}_t^k,\! \rho_{\text{overflow}},\! \mathbb{I}_{\text{border}}^k \!\right] \!\!\in\!\! \mathbb{R}^{6 + 2(m\!-\!1)},
\end{equation}
where $\{{\Delta \mathbf{p}_j^k\}_{j \neq k}}=\frac{\mathbf{p}_t^j - \mathbf{p}_t^k}{\|\mathbf{b}\|}$ denotes the relative positions of other AUVs, with $\mathbf{p}_t^k \in \mathbb{R}^2$ denoting the 2D position of the $k$-th AUV, and $\mathbf{b} = [X_{\text{max}}, Y_{\text{max}}]^\top$ defining the environmental boundaries. 
Besides, $\Delta \mathbf{p}_{\text{target}}^k = \frac{\mathbf{p}_{\text{target}}^k - \mathbf{p}_t^k}{\|\mathbf{b}\|}$ denotes the relative position to the target SN, with $\mathbf{p}_{\text{target}}^k$ representing the position of the target SN assigned to the $k$-th AUV. 
The normalized self-position $\tilde{\mathbf{p}}_t^k = \frac{\mathbf{p}_t^k}{\|\mathbf{b}\|}$ encodes spatial awareness within the bounded domain. 
Moreover, $\rho_{\text{overflow}} = \frac{N_{\text{DO}}}{N_{\text{POI}}}$ represents the data overflow ratio, a global congestion indicator reflecting the proportion of sensor nodes whose data buffers are full at time $t$. 
A higher $\rho_{\text{overflow}}$ implies severe network congestion and guides AUVs to prioritize heavily loaded nodes for data collection. 
In contrast, $\mathbb{I}_{\text{border}}^k \in \{0, 1\}$ is the border violation flag, which takes the value $1$ when the predicted next position $\tilde{\mathbf{p}}_{t+1}^{k} = \mathbf{p}_t^{k} + v_t^{k} \Delta t [\cos(\theta_t^{k}), \sin(\theta_t^{k})]^\top$ lies outside the operational boundary $\mathcal{Q} = [0, X_{\max}] \times [0, Y_{\max}]$. 
Including both quantities in the state vector allows the agent to simultaneously perceive global network congestion and local safety constraints, which are essential for maintaining balanced and efficient underwater operations.

The full action vector $a_t^k \in \mathcal{A}$ for the $k$-th AUV is a 2D continuous vector:
\begin{equation}
    a_t^k \!=\! [v_{\text{norm}}, \theta_{\text{norm}}]^\top, \quad v_{\text{norm}} \!\in\! [-1, 1], \ \theta_{\text{norm}} \!\in\! [-1, 1],
\end{equation}
where $v_{\text{norm}}$ denotes the expected normalized speed, which is linearly mapped to the physical range $[V_{\text{min}}, V_{\text{max}}]$:
$v_{\text{norm}}=\frac{2(v_t^k-V_{\text{min}})}{V_{\text{max}} - V_{\text{min}}}-1$. 
$\theta_{\text{norm}}$ stands for the expected normalized direction and is scaled to $[-\pi, \pi]$: $\theta_{\text{norm}}=\frac {\theta_t^k}{\pi}$.

\begin{figure*}[!t]
    \centering
    \subfigure[Sea conditions at 25s]{\includegraphics[width=0.325\linewidth]{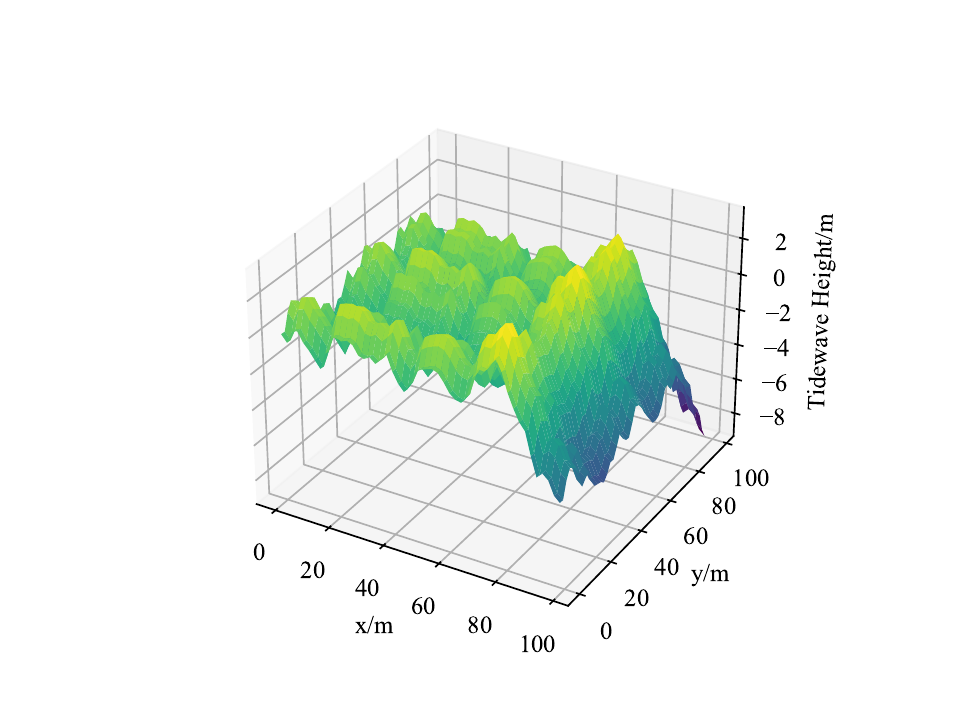}}   
    \subfigure[Sea conditions at 50s]{    \includegraphics[width=0.325\linewidth]{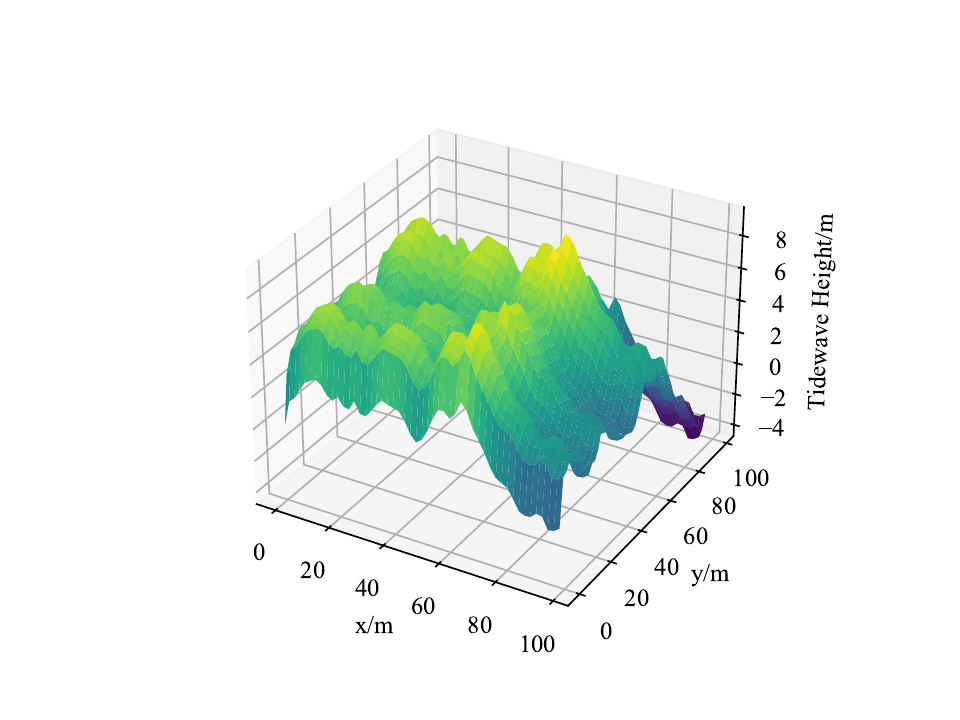}}  
    \subfigure[Sea conditions at 75s]{\includegraphics[width=0.325\linewidth]{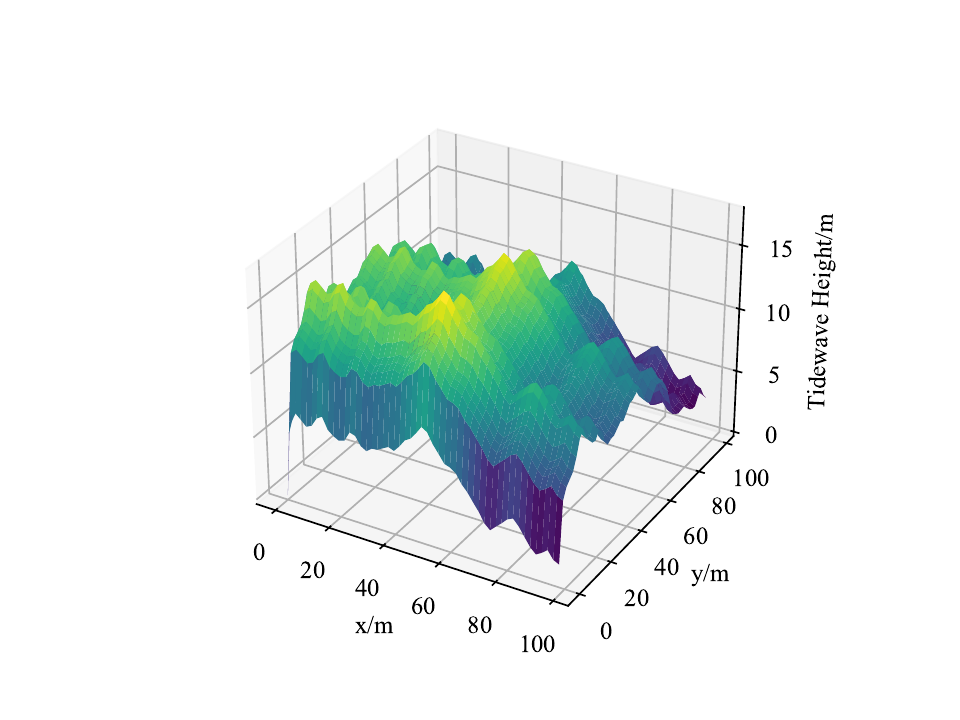}}
    \caption{The illustration of realistic extreme sea condition simulation via finite difference method at three different time points. (a) Sea conditions at 25s. (b) Sea conditions at 50s. (c) Sea conditions at 75s.}
    \label{fig3}
\end{figure*}

    \begin{figure*}[!t]
        \centering
        \subfigure[Sum data rate \qquad \qquad \qquad \qquad \qquad \qquad (b) Energy consumption \qquad \qquad \qquad \qquad \qquad  (c) Average reward per timestep \!\!\!\!\!\!\!\!\!\!\!\!\!\!\!\!\!\!\!\!\!\!\!\!\!\!\!\!\!\!\!\! ]{\includegraphics[width=1.0\linewidth]{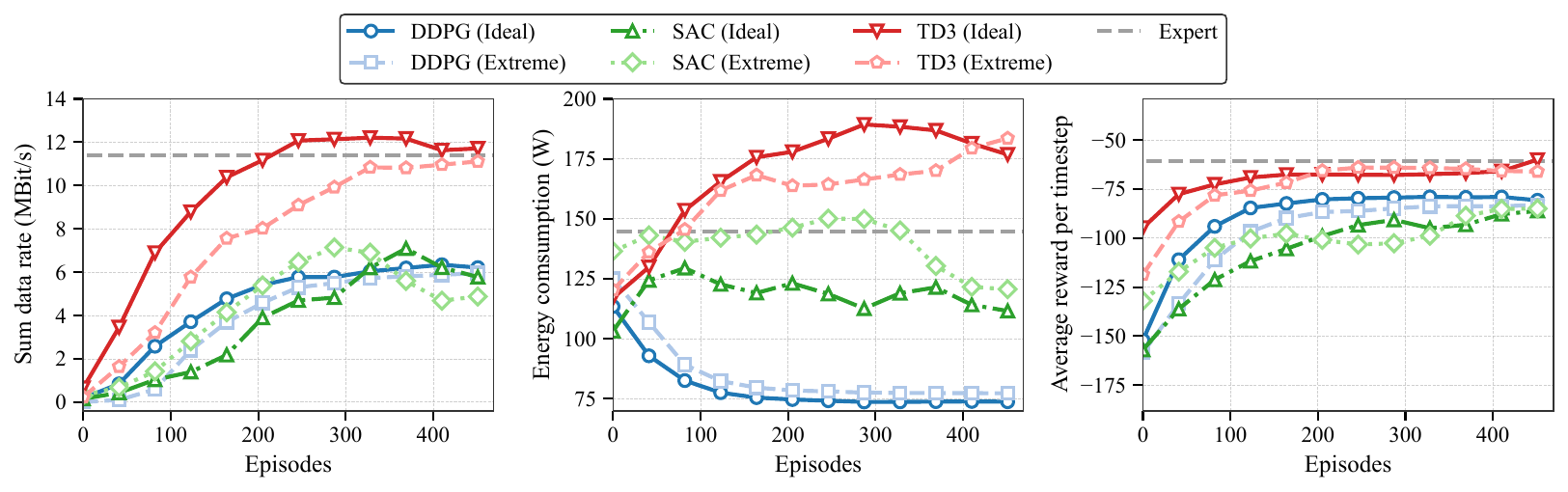}}
        
        \caption{The curves of sum data rate, energy consumption and average reward per timestep, using TD3, DDPG and SAC for RL training in ideal and extreme sea conditions, respectively. (a) Sum data rate. (b) Energy consumption. (c) Average reward per timestep.}
        \label{fig4}
        \end{figure*}

\begin{table*}[!ht]
            \centering
            \caption{Comparison of Different RL Algorithms in Ideal and Extreme Sea Conditions}
            \begin{tabular}{lcccccc}
            \toprule
              & SDR (Mbps) & EC (W) & ARPT & SSN & Trajectory Lengths (m)\\
            \midrule
            TD3 (ISC) &11.20±1.93 &138.19±3.80&-60.69±7.81&47.0±5.52& (1354±75,1340±45)\\
            TD3 (ESC)  &10.78±1.70& 147.81±5.09&-62.95±8.60 & 45.4±6.81 & (1389±38,1371±34)\\
            DDPG (ISC)         & 6.63±0.97 & 73.72±1.11 & -80.70±6.94 & 30.8±3.97 & (1138±41,1089±63)\\
            DDPG (ESC)      & 6.35±1.57  & 77.23±0.89  & -82.28±7.41 & 29.6±4.91  & (1108±34,1095±42)\\
            SAC (ISC)  & 5.97±2.66   & 111.57±13.23  & -87.86±13.10 & 27.2±7.35  & (1156±89,1171±78)\\
            SAC (ESC)    & 5.58±2.35  & 117.76±17.30 & -89.58±15.22 & 25.0±6.62 & (1132±94,1109±62)\\
            Baseline [5]  & 5.93±3.51  & 112.30±10.56 & -87.95±12.08 & 25.8±7.94 & (1170±71,1196±77)\\
            \bottomrule
            \end{tabular}
        \end{table*}

Since the reward plays a critical role in RL training, we design a shaped reward function that balances multiple objectives:
\begin{equation}
\mathcal{R} = \boldsymbol{\lambda}^T \boldsymbol{R} = \sum_{i=1}^p \lambda_i R_{i},
\end{equation} 
where $\boldsymbol{R} = \{R_{1}, R_{2}, \ldots, R_{p}\}$ represents distinct objectives, and $\boldsymbol{\lambda} = \{\lambda_1, \lambda_2, \ldots, \lambda_p\}$ controls the relative weighting of these objectives. 

Substituting into the underwater data collection task, our reward function is designed as follows:
\begin{equation}
\begin{split}
\mathcal{R} =  \underbrace{-0.6 d_k^{\text{target}} - 0.05 N_{\text{DO}}+ 12 \mathbb{I}_{\text{TL}}}_{\text{Approaching target SNs}}- \underbrace{0.085 E_k}_{\text{Energy consumption}}   \\ 
 \underbrace{-\sum_{j \neq k} 6 (12 - d_{kj}) - 0.1\mathbb{I}_{\text{border}}}_{\text{Safety requirements}},
\end{split}
\end{equation}
where $d_k^{\text{target}}$ denotes the AUV–SN distance, $\mathbb{I}_{\text{border}}$ and $N_{\text{DO}}$ correspond to the border violation and network congestion penalties, $d_{kj}$ is the inter-AUV distance, $\mathbb{I}_{\text{TL}}$ represents the transmission success flag, and $E_k$ indicates energy consumption. 
Accordingly, $N_{\text{DO}}$ and $\mathbb{I}_{\text{border}}^k$ jointly influence the RL mechanism: the former informs the policy regarding the overall data backlog, while the latter constrains unsafe maneuvers near domain boundaries. 
Together, they allow the policy to learn a cooperative behavior that enhances both communication throughput and navigation safety.

\begin{figure*}[!t]
 \centering
 \subfigure[Sum data rate]{\centering\includegraphics[width=0.328\linewidth]{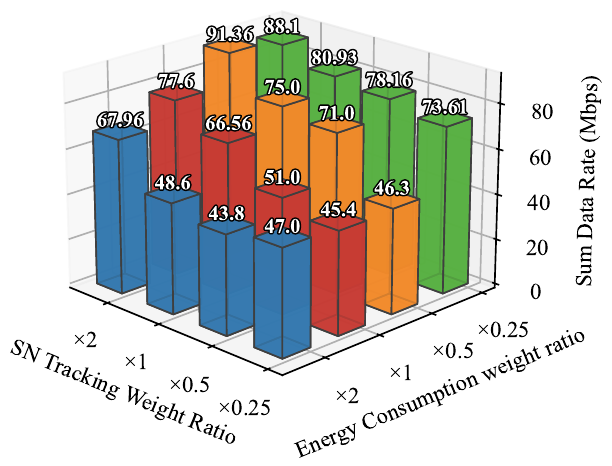}}
    \subfigure[Energy consumption]{\centering\includegraphics[width=0.328\linewidth]{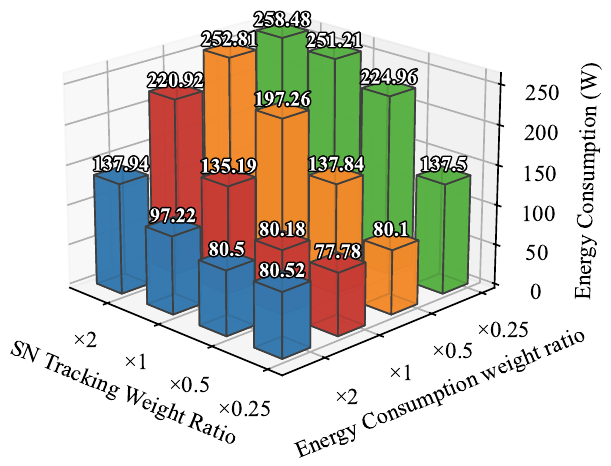}}
    \subfigure[Serviced sensor nodes]{\centering\includegraphics[width=0.328\linewidth]{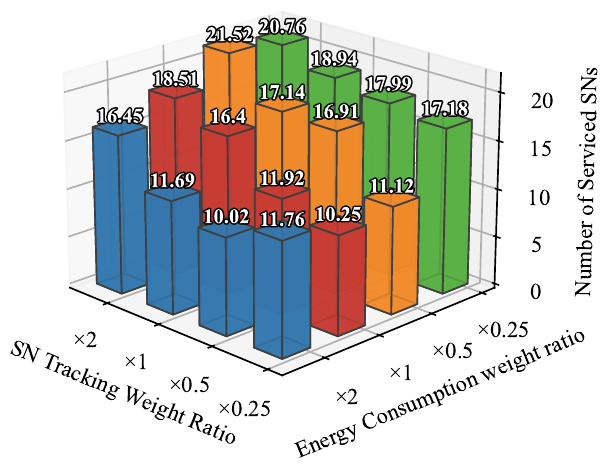}}
  \caption{Performance comparison under different weighting ratios between energy- and SN-related reward terms. The results illustrate the trade-off between energy efficiency and data collection capability when scaling the relative weights of these two terms. (a) Sum data rate. (b) Energy consumption. (c) Serviced sensor nodes.}
  \end{figure*}

\begin{figure*}[!t]
 \centering
 \subfigure[Safety violation times]{\centering\includegraphics[width=0.24\linewidth]{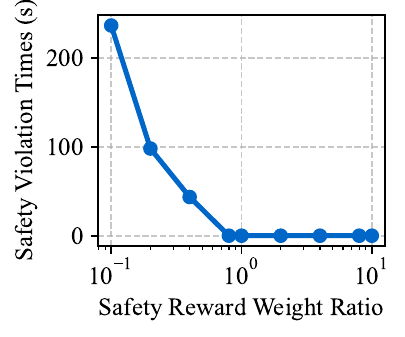}}
 \subfigure[Sum data rate]{\centering\includegraphics[width=0.24\linewidth]{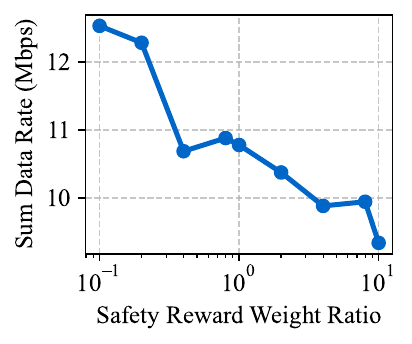}}
    \subfigure[Energy consumption.]{\centering\includegraphics[width=0.24\linewidth]{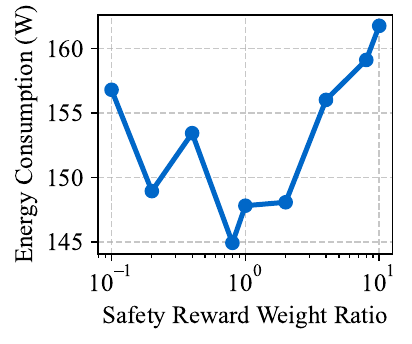}}
    \subfigure[Serviced Sensor Nodes.]{\centering\includegraphics[width=0.24\linewidth]{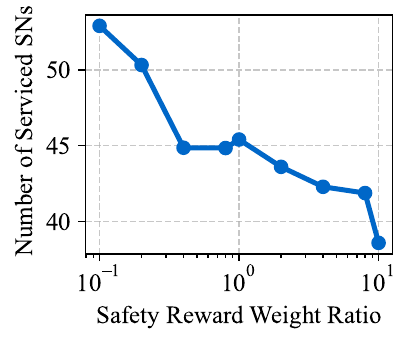}}
  \caption{Performance trends under different safety-related reward coefficients. The plots show the sensitivity of key performance metrics to the scaling of the safety weight. (a) Safety violation times. (b) Sum data rate. (c) Energy consumption. (d) Serviced Sensor Nodes.}
  \end{figure*}
    
\subsection{Experiment Results and Analysis} 
To comprehensively evaluate the feasibility and effectiveness of the proposed system, we conducted a series of systematic experiments involving both training and evaluation phases, where three state-of-the-art RL algorithms: TD3, DDPG, and Soft Actor-Critic (SAC) were implemented to train two AUVs coordinated by a USV under both ideal and extreme sea conditions. As shown in Fig.~4, the RL training process exhibits stable convergence across random seeds, and the system ultimately reaches an expert-level policy, indicating consistent and robust learning behavior. Here, the convergence is declared only when the 25-episode moving averages of these metrics exhibit slope magnitudes below  0.2 for at least 50 consecutive episodes and no further improvement is observed in validation metrics, ensuring that the selected checkpoint corresponds to a fully converged rather than prematurely stopped policy. Therefore, the expert policy here refers to the fully converged model that achieves the highest validation performance across these metrics. Quantitatively, the Sum Data Rate (SDR) of TD3 increases rapidly and stabilizes around 11~Mbps, approximately 70\% higher than that of DDPG and SAC ($\approx6$~Mbps) under ideal conditions. In contrast, although TD3 incurs a slightly higher peak Energy Consumption (EC), which is roughly 20\% higher than SAC, it maintains the highest Average Reward Per Timestep (ARPT) ($\approx-60$), showing a 25–30\% improvement over both DDPG and SAC in long-term task performance. Taken together, these quantitative comparisons confirm that TD3 not only converges stably to a rigorously defined expert policy but also achieves the best trade-off between data collection efficiency and energy cost, significantly outperforming DDPG and SAC across key evaluation metrics.

In addition, to further assess the robustness of the proposed system, we conducted extensive environmental generalization experiments comparing performance under both Ideal Sea Conditions (ISC) and Extreme Sea Conditions (ESC). The results, summarized in Table III (where SSN denotes Serviced Sensor Nodes) show that, despite the challenging environmental factors in ESC, the system maintains performance levels comparable to those achieved under ISC. This consistent performance across different environmental scenarios highlights the system’s strong robustness and adaptability in extreme sea conditions. 

Besides, we further introduced the “total distance traveled per AUV” (denoted as trajectory lengths in the tables) as an additional performance metric to complement these evaluations and provide a more comprehensive understanding of cooperative behavior and task assignment. As shown in Table III, the trajectory lengths of individual AUVs under both ISC and ESC remain highly consistent, for example, (1354 ± 75 m, 1340 ± 45 m) for TD3 (ISC) and (1389 ± 38 m, 1371 ± 34 m) for TD3 (ESC), with a variance below 2\%. This quantitative stability demonstrates that the proposed collaborative system effectively balances workload distribution among AUVs, preventing any single vehicle from being overburdened. Moreover, compared to baseline methods, the utilized approach (TD3) achieves an average increase of about 80\% in served SNs with only approximately an 20\% increase in trajectory lengths, confirming the system’s superiority in path-planning efficiency and energy conservation.

  \begin{figure*}[!t]
        \centering
        \subfigure[Average reward per timestep]{
        \includegraphics[width=0.475\linewidth]{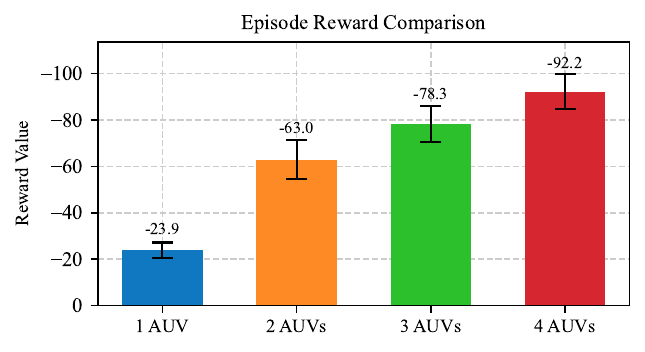}}
        \subfigure[Sum data rate]{
        \includegraphics[width=0.475\linewidth]{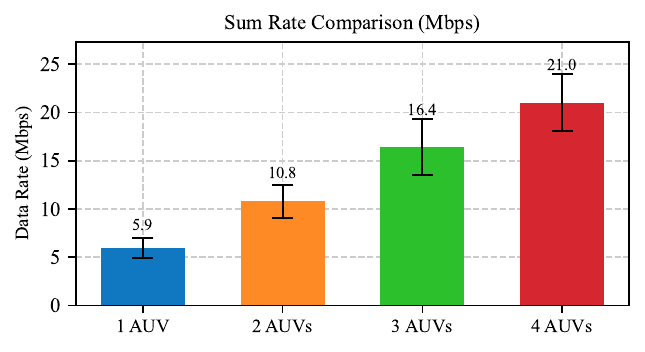}}
        \subfigure[Energy consumption]{
        \includegraphics[width=0.475\linewidth]{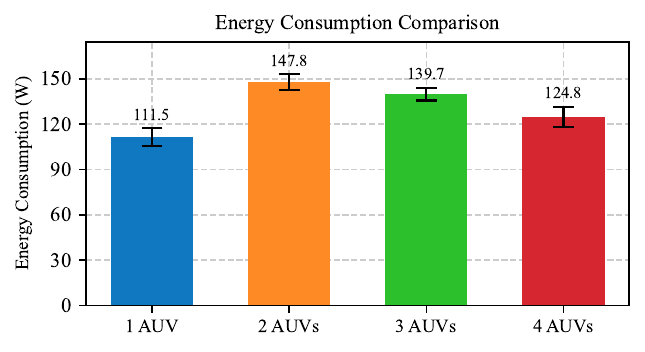}}
        \subfigure[Serviced Sensor Nodes]{
        \includegraphics[width=0.475\linewidth]{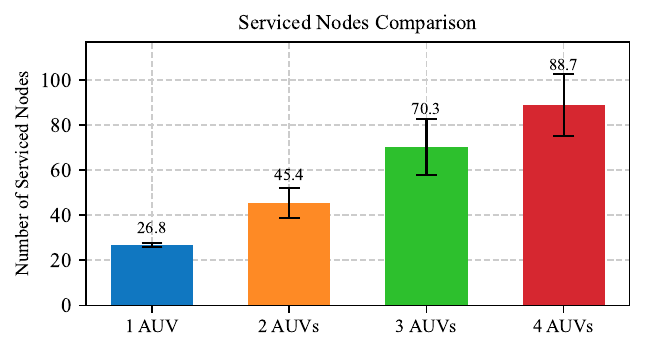}}
        \caption{The histogram of average reward per timestep, sum data rate, energy consumption and average reward per timestep, using RL to train 1 AUV, 2 AUVs, 3 AUVs and 4 AUVs in extreme sea conditions, respectively. (a) Average reward per timestep. (b) Sum data rate. (c) Energy consumption. (d) Serviced sensor nodes.}
        \label{fig6}
        \end{figure*}

\begin{table*}[!ht]
            \centering
            \caption{Comprehensive Performance Comparison under Different Numbers of AUVs}
            \begin{tabular}{lccccc}
            \toprule
              & SDR (Mbps) & EC (W) & ARPT & SSN & Trajectory Lengths (m) \\
            \midrule
            $N=1$ &5.94±1.03 &111.47±5.79&-23.87±3.27&26.8±1.04& (1309±91)\\
            $N=2$  &10.78±1.70& 147.81±5.09&-62.95±8.60 & 45.4±6.81& (1389±38,1371±34)\\
            $N=3$   & 16.39±2.90 & 139.72±4.18 & -78.26±7.84 & 70.3±12.45& (1418±108,1452±72,1380±71) \\
            $N=4$ & 21.02±2.94  & 124.83±6.63  & -92.21±7.51 & 88.7±13.74& (1445±125,1407±159,1512±153,1420±86)  \\
            \bottomrule
            \end{tabular}
        \end{table*} 

Building upon the above experimental evaluations, we further delve into the design of the reward function and the determination of a robust set of reward weights. As shown in Eq. 24, the reward integrates interpretable terms for target-tracking ($d_k^{\text{target}}$), safety ($\mathbb{I}_{\text{border}}$ and $d_{kj}$), data overflow and transmission success ($N_{\text{DO}}$ and $\mathbb{I}_{\text{TL}}$), and energy consumption ($E_k$). Correspondingly, we analyzed how the relative weighting of these factors affects system behavior. Since safety is fundamental for multi-AUV coordination, safety-related terms were prioritized, which effectively reduced boundary violations and collision risks to negligible levels. After ensuring stable safe behaviors, we examined the trade-off between data-collection efficiency (affected by $d_k^{\text{target}}$, $N_{\text{DO}}$, $\mathbb{I}_{\text{TL}}$) and energy usage ($E_k$), revealing a Pareto relationship between operational gain and energy expenditure.

\begin{figure*}[!t]
    \centering
    \subfigure[2 AUVs]{\includegraphics[width=0.325\linewidth]{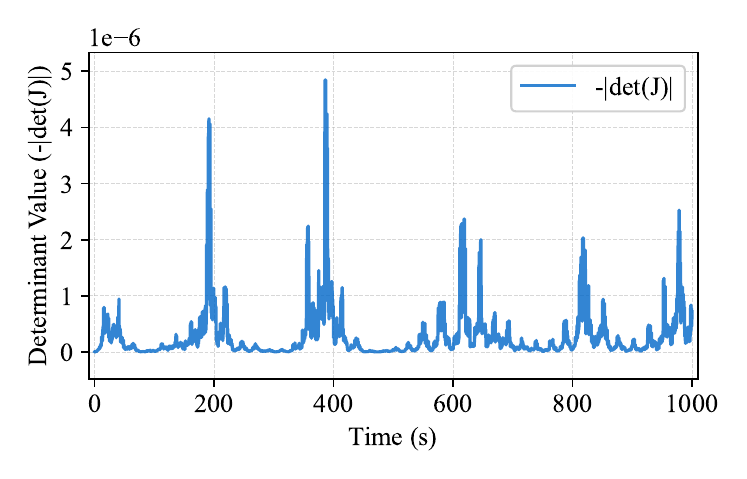}}   \subfigure[3 AUVs]{    \includegraphics[width=0.325\linewidth]{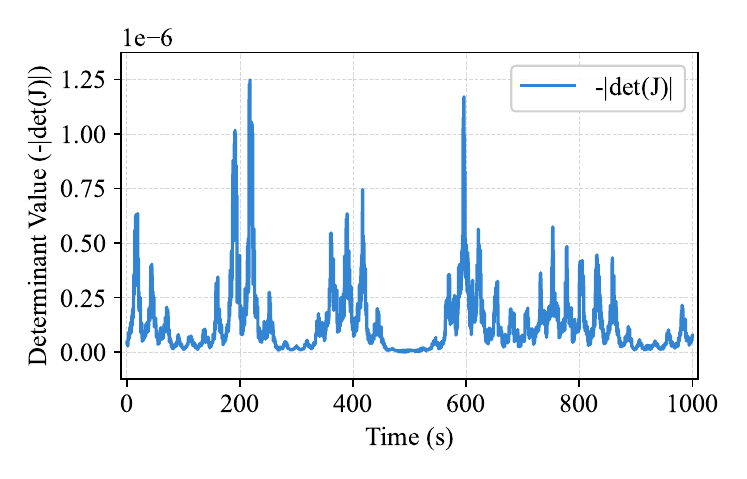}}  
    \subfigure[4 AUVs]{\includegraphics[width=0.325\linewidth]{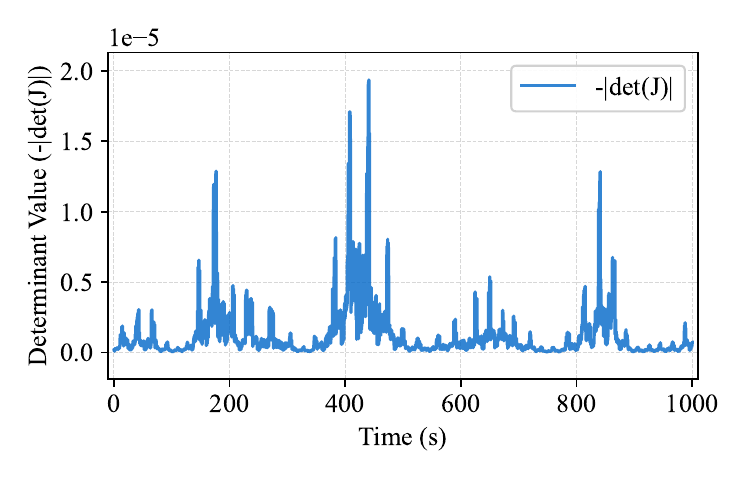}}
    \caption{Negative determinant value evolution of FIM in an operational episode using different numbers of AUVs. (a) 2 AUVs. (b) 3 AUVs. (c) 4 AUVs.}
    \label{fig2}
\end{figure*}

\begin{figure*}[!t]
    \centering
    \subfigure[2 AUVs]{   \includegraphics[width=0.33\linewidth]{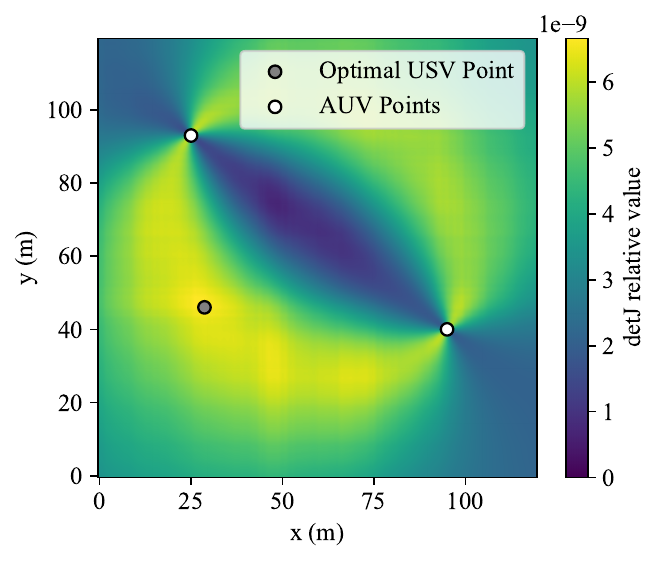}}\!\!\!
    \subfigure[3 AUVs]{   \includegraphics[width=0.33\linewidth]{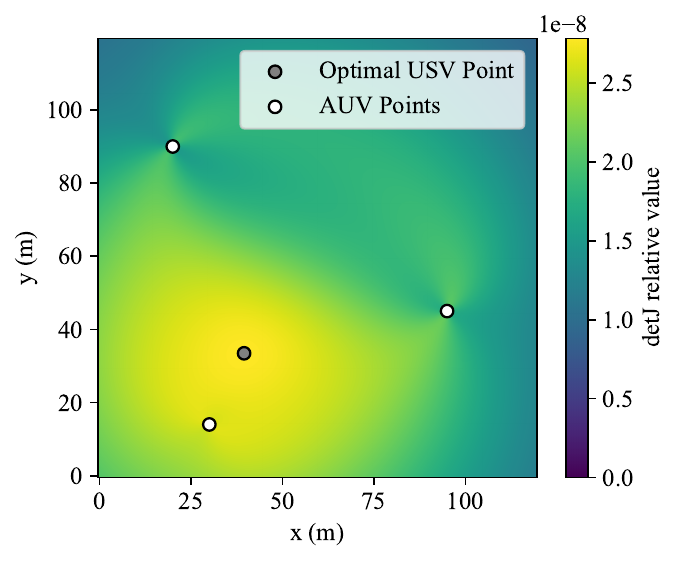}}\!\!\!
    \subfigure[4 AUVs]{\includegraphics[width=0.33\linewidth]{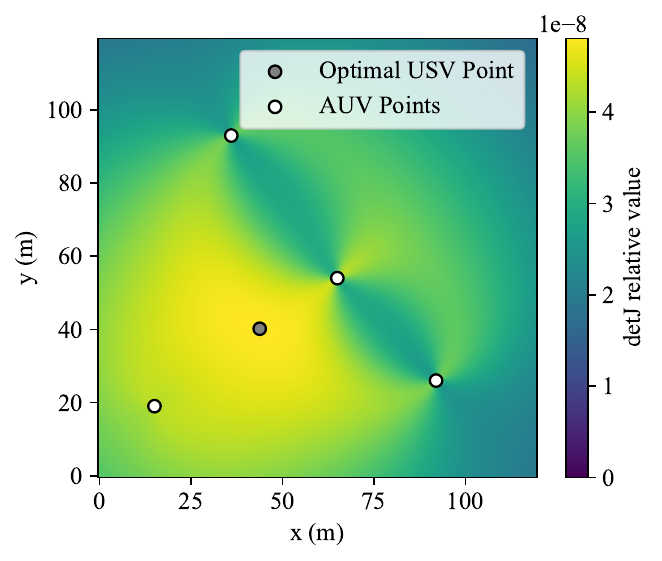}}
    \caption{The illustrations of the USV-AUV coordinate relationships in the collaborative system at some point. The optimal USV positions were determined through real-time optimization of FIM determinant value to minimize AUV positioning errors. (a) 2 AUVs: The optimal USV position is at coordinates (28.67~m, 46.01~m). (b) 3 AUVs: The optimal USV position is at coordinates (39.50~m, 33.45~m). (c) 4 AUVs: The optimal USV position is at coordinates (43.77~m, 40.17~m).}
    \label{fig5}
\end{figure*}

\begin{figure*}[!t]
        \centering
        
        \subfigure[Time period 1 \qquad \qquad  \qquad \qquad \qquad \qquad \qquad (b) Time period 2 \qquad \qquad \qquad \qquad \qquad \qquad \qquad  (c) Time period 3 \!\!\!\!\!\!\!\!\!\!\!\!\!\!\!\!\!\!\!\!\!\!\!\!\!\!\!\!\!\!\!\! ]{
        \includegraphics[width=1.0\linewidth]{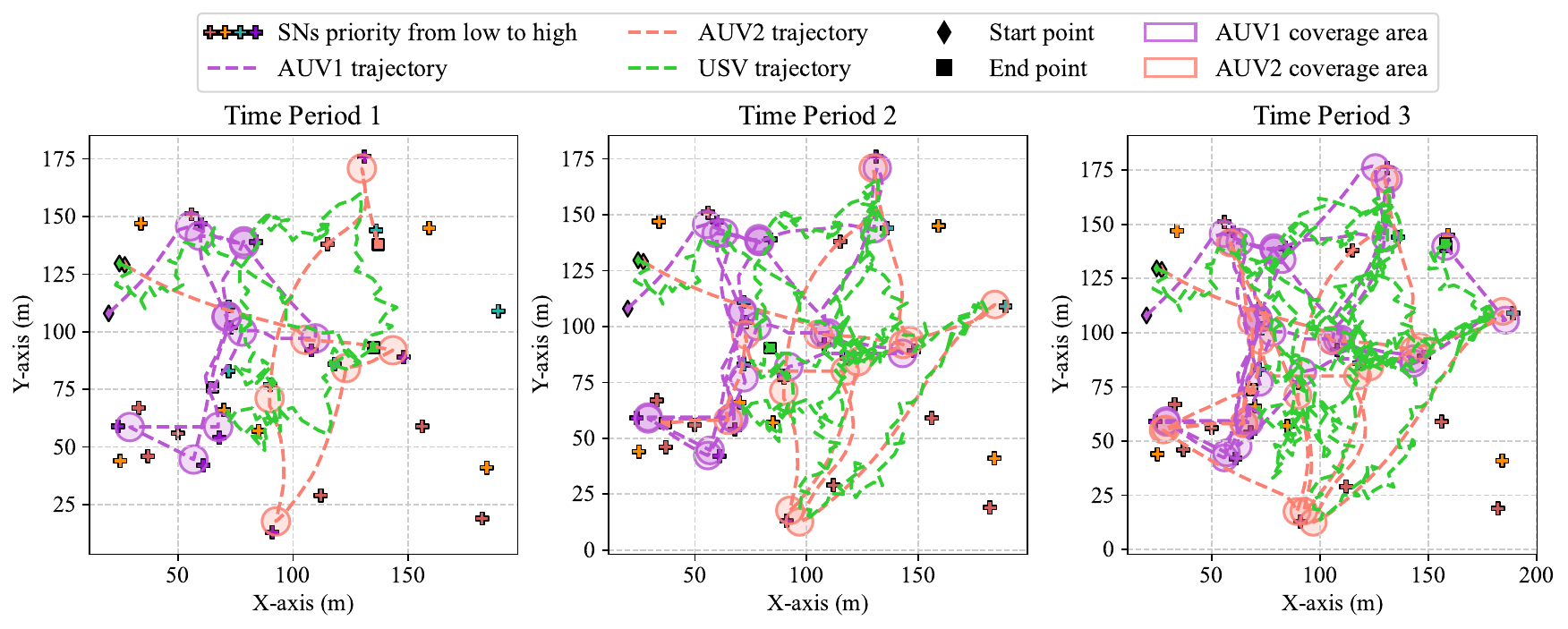}}

        \caption{Trajectories of AUVs and USV in three different time periods of an operational episode. (a) Time period 1. (b) Time period 2. (c) Time period 3.}
\end{figure*}

\begin{figure}[!t]
        \centering
        \includegraphics[width=1.0\linewidth]{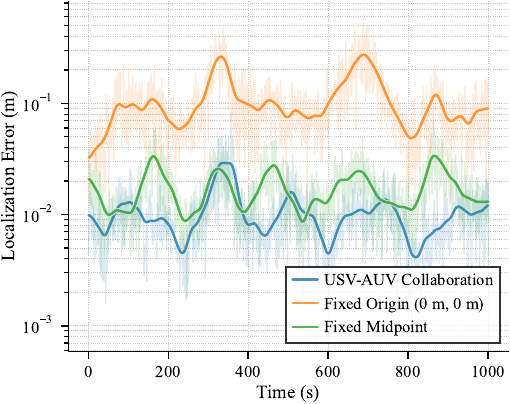}

        \caption{Positioning error of the AUVs with USV fixed at (0,0) and (100, 100), and path planning using FIM optimization, respectively.}
\end{figure}

\begin{table*}[!ht]
    \centering
    \caption{Comparison of Major Performance Metrics under Different USV Path Planning Policies}
    \begin{tabular}{lccccc}
    \toprule
      & SDR (Mbps) & EC (W) & ARPT & SSN \\
    \midrule
    \textbf{TD3 (ESC)} &\textbf{10.78±1.70}& \textbf{147.81±5.09}&\textbf{-62.95±8.60} & \textbf{45.4±6.81}\\
    TD3-R/S (ESC)&  9.28±2.74 & 154.14±9.40 & -65.89±9.45 & 39.0±10.35  \\
    TD3-R/D (ESC) & 10.47±2.23 & 150.37±9.51 & -64.38±8.25 & 44.9±8.41  \\
    \bottomrule
    \end{tabular}
\end{table*}

\begin{table*}[!ht]
    \centering
    \caption{Comparison of Major Performance Metrics under Varying Numbers of FIM Optimization Iterations}
    \label{table:2}
    \begin{tabular}{lccccc}
    \toprule
     Iteration counts & SDR (Mbps) & EC (W) & ARPT & SSN \\
    \midrule
    \verb|nit=6| & 8.67±2.29 & 156.31±9.23 & -67.51±10.32 & 37.3±7.93\\
    \verb|nit=12|  & 10.08±1.66 & 155.45±6.25 & -65.64±6.30 & 43.0±6.82\\
    \verb|nit=18|  & 10.40±1.88 & 149.92±9.11 & -63.81±7.56 & 44.5±6.68 \\
    \texttt{\textbf{nit=24}} & \textbf{10.82±2.10} & \textbf{152.17±7.56} & \textbf{-62.09±8.81} & \textbf{46.3±7.57}  \\
    \bottomrule
    \end{tabular}
\end{table*}

\begin{figure*}[!ht]
 \centering
 \subfigure[Positioning error of AUVs under different USV path planning policies]{\centering\includegraphics[width=0.455\linewidth]{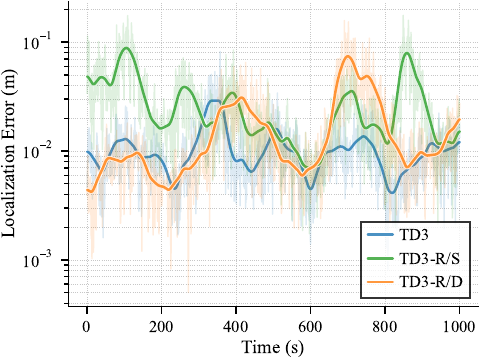}}\quad
    \subfigure[Positioning error of AUVs under different FIM optimization iterations]{\centering\includegraphics[width=0.455\linewidth]{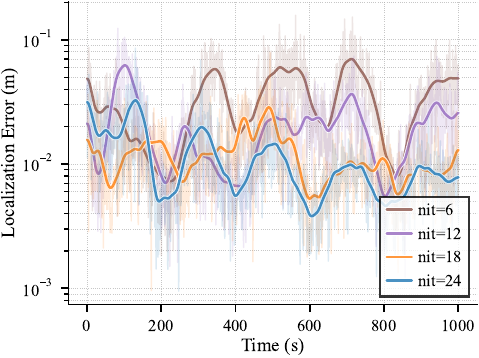}}
  \caption{\small AUV positioning error under different USV path planning policies and different number of FIM optimization iterations. (a) Positioning error of AUVs under different USV path planning policies. (b) Positioning error of AUVs under different FIM optimization iterations. }
  \end{figure*}
  
We first investigate how the relative weighting between the other two reward components influences task performance. To this end, we scaled their weights by factors of $(0.25, 0.5, 1, 2)$. Controlled experimental results in Fig.~5 showed that overly small energy penalties led to high-speed but energy-intensive trajectories, whereas excessive penalization limited mobility and reduced the SSNs; moderate weighting achieved an optimal balance across the SDR, EC, and SSN. Subsequently, we multiply the safety-related term by scaling factors ranging from 0.1 to 10 and analyze how safety performance (i.e., the total violation times of all AUVs) and other major metrics evolve. As depicted in Fig. 6, further tuning of the safety coefficient demonstrated that increasing its weight rapidly suppressed violations near a ratio of one, beyond which improvements saturated and minor performance degradation appeared. Based on these observations, we selected a safety weight near this knee point and balanced SN-tracking and energy weights around unity, yielding the final configuration that ensures stable convergence and strong generalization in both ideal and extreme sea conditions.

Based on the above results, we further evaluated the scalability and coordination efficiency of the proposed USV–AUV collaborative system by testing configurations with 1, 2, 3, and 4 AUVs under extreme sea conditions. As summarized in Table IV and illustrated in Fig. 7, the system exhibits clear and consistent performance trends as the team size increases. Although ARPT shows a gradual decline due to higher system complexity, both SDR and SSN improve markedly: SDR increases from 5.9~Mbps to 21.0~Mbps, and SSN increases from 26.8 to 88.7 as the number of AUVs grows from 1 to 4. This pattern arises from the saturation of discrete positive rewards and the accumulation of energy, waiting, and proximity penalties that scale with fleet size. Hence, the declining ARPT reflects denser penalty events rather than degraded performance. Mission-level metrics (SDR and SSN) confirm that overall efficiency and scalability remain robust, demonstrating that multi-AUV cooperation effectively distributes task loads and enhances overall operational efficiency through parallelized data collection. Meanwhile, the trajectory lengths remain highly consistent among all AUVs across different configurations; for example, they are (1389 ± 38~m, 1371 ± 34~m) for $N = 2$ and (1445 ± 125~m, 1407 ± 159~m, 1512 ± 153~m, 1420 ± 86~m) for $N = 4$, indicating that the cooperative RL framework achieves balanced workload allocation without overburdening any single vehicle. Although larger teams naturally entail higher energy consumption, the increase remains moderate; for instance, the EC for four AUVs is only about 1\% higher than that for three AUVs and even 10\% lower than that for a single AUV, confirming that the proposed framework can scale efficiently, optimize path planning and resource utilization, and maintain robust, energy-aware coordination even under extreme sea conditions.

To further evaluate the robustness and scalability of the proposed USV-AUV collaborative system, we conducted additional experiments employing 2, 3, and 4 AUVs for underwater data collection tasks in extreme sea conditions. A key focus of these experiments was to minimize the negative determinant of the FIM in real-time during each operational episode, thereby optimizing USV path planning for enhanced AUV positioning accuracy. The results demonstrate that, despite significant environmental disturbances (including ocean turbulence and strong waves), the system consistently maintained the FIM determinant's negative value within a minimized range throughout each operational episode. As shown in Fig. 8, this achievement was observed across all tested configurations (2, 3, and 4 AUVs), with the minimization performance demonstrating particular robustness in extreme sea conditions. Furthermore, through visualization in Fig. 9, it was clearly observed that the USV's path planning points consistently fell in the region with the lowest AUV positioning error. The optimal USV positions in the USV-AUV collaborative system were determined as follows:
\begin{itemize}
    \item 2 AUVs configuration: Optimal USV position at (28.67~m, 46.01~m) for AUVs located at (25~m, 93~m) and (95~m, 40~m).

    \item 3 AUVs configuration: Optimal USV position at (39.50~m, 33.45~m) for AUVs located at (20~m, 90~m), (95~m, 45~m), and (30~m, 14~m).

    \item 4 AUVs configuration: Optimal USV position at (43.77~m, 40.17~m) for AUVs located at (36~m, 93~m), (65~m, 54~m), (15~m, 19~m), and (92~m, 26~m).
\end{itemize}

These findings demonstrate the system's capability to dynamically adapt the USV's position in response to real-time AUV localization requirements. Notably, when the USV positioned itself at the point of minimized FIM determinant negative value, we observed a significant reduction in AUV positioning errors. This empirical evidence strongly supports the fundamental importance of FIM optimization in effective path planning for our proposed USV-AUV collaborative system.

In addition to the base experiments, we leveraged the expert policy trained using the TD3 algorithm to coordinate a multi-AUV system executing the underwater data collection task in extreme sea conditions. Fig. 10 illustrates the complete trajectories of both the AUVs and the USV during a representative RL training episode, showcasing their coordinated movement patterns and task execution strategies. To quantitatively evaluate the advantages of our USV-AUV collaborative framework, we conducted a detailed analysis of the positioning accuracy under three distinct operational scenarios: 

\begin{itemize}
\item Employing USV path planning through FIM optimization techniques.
\item Maintaining the USV at a fixed position at the original coordinates (0~m,0~m). 
\item Maintaining the USV at a fixed position at midpoint coordinates (100~m,100~m). 
\end{itemize}

The positioning error comparison presented in Fig. 11 clearly demonstrates that the FIM-optimized USV path planning approach achieves significantly lower positioning errors compared to the fixed-position alternatives. These results provide compelling evidence for the superior performance of USBL positioning when combined with optimized USV path planning, even when operating under the most challenging sea conditions. The system's ability to maintain high positioning accuracy in such environments further confirms its practical applicability for real-world underwater operations.

Moreover, to quantitatively evaluate the performance and robustness of the proposed FIM-based USV path planning module against alternative RL-based planning strategies commonly used in similar underwater cooperative settings, we designed two comparative baselines in which the USV’s trajectory was determined solely by standalone RL policies using the TD3 algorithm. The first variant (TD3-R/S) employed a sparse reward emphasizing the coverage of target sensor nodes, while the second (TD3-R/D) adopted a dense reward derived from the Signal-to-Noise Ratio (SNR) of the acoustic signals received by the USV from the AUVs. To further assess robustness under computational constraints, we analyzed how reducing the number of iterations in the FIM determinant maximization process influences localization accuracy and planning stability. The quantitative comparisons summarized in Table~V and Table~VI evaluate SDR, EC, ARPT, and SSN, while Fig.~11 illustrates the evolution of AUV positioning errors under different conditions. As shown in Table~V, although TD3-R/D slightly outperforms TD3-R/S in SDR and SSN, both RL-based baselines still underperform the proposed FIM-optimized planner, which achieves the highest SDR (10.78 Mbps) and lowest energy consumption (147.81 W). This demonstrates that the explicit incorporation of information-theoretic guidance significantly enhances data transmission efficiency, control smoothness, and overall trajectory stability.

Finally, we conduct a systematic computational complexity analysis. In our implementation, the FIM-based optimization employs the differential evolution algorithm \cite{43} provided by the scipy library \cite{44}, with the number of optimization iterations (denoted by the attribute \texttt{\textbf{nit}} in scipy) used as the primary evaluation metric. Table~VI reveals that, as the number of FIM optimization iterations increases, system performance consistently improves across all evaluated metrics. Convergence trends emerge beyond 18 iterations, and optimal results are attained at 24 iterations, where the Spatial Data Rate (SDR) and the Signal-to-Noise Score (SSN) reach 10.82~Mbps and 46.3, respectively. Under this setting, in which 24 iterations are executed on a single CPU core with a fixed clock frequency of 3.6~GHz, representative of typical single-core laptop performance around 2021, the average runtime per optimization is 61~ms. This result, obtained under relatively conservative computational conditions, strongly suggests that real-time execution of the FIM optimization is highly feasible on modern multi-core systems. These numerical trends are further corroborated by Fig.~11, which shows that the localization error of the proposed FIM-based planner remains consistently lower and exhibits reduced temporal fluctuations compared to RL-only baselines. Moreover, increasing the number of optimization iterations progressively suppresses high-frequency deviations in the localization trajectory. Collectively, these results confirm that the FIM-guided optimization mechanism provides superior localization precision, energy efficiency, and stability, demonstrating the reliability and adaptability of the proposed USV–AUV collaborative framework. The observed performance gains clearly stem from the mathematical coupling between information-driven trajectory optimization and RL-based decision-making, validating the effectiveness and generalizability of the proposed bi-level information-structured design.

\section{Conclusion} 
In this study, we developed a USV–AUV collaborative system designed to enhance underwater task performance under extreme sea conditions. The framework comprises two core components: (1) high-precision multi-AUV localization, achieved through FIM-based USV path planning, and (2) RL-driven multi-AUV cooperative task execution. Extensive experiments on underwater data collection tasks verified the system’s feasibility and demonstrated its superior coordination efficiency, robustness, and adaptability in challenging marine environments. To facilitate further research and practical development in this field, we have open-sourced the simulation framework at https://github.com/360ZMEM/USV-AUV-colab
.

In future work, we aim to extend the proposed USV–AUV collaborative framework from high-fidelity simulation to real-world deployment, addressing the Sim-to-Real gap and enabling online learning and adaptive control for operation under extreme sea conditions. The system, integrating an information-theoretic FIM-based USV planner and a physically interpretable RL controller, will be enhanced with LLM-driven adaptive mechanisms that dynamically adjust reward weights and control parameters based on environmental feedback, improving robustness and generalization in uncertain marine environments. On the experimental side, we are assembling two customized BlueROV2-based AUVs and a USV platform with embedded processors (e.g., NVIDIA Jetson, ARM modules) for real-time inference and communication-efficient coordination. A four-stage validation roadmap, which covers pool, harbor, nearshore, and offshore trials, will progressively assess waterproofing, localization accuracy, communication stability, and control performance, ultimately bridging the gap between simulation and scalable ocean deployment.


\bibliographystyle{IEEEtran}
\bibliography{IEEEexample}

\addtolength{\textheight}{-12cm}

\end{document}